\pdfoutput=1

\documentclass[11pt]{article}
\usepackage{emnlp2021}
\usepackage{times}
\usepackage{latexsym}
\usepackage{graphicx}
\usepackage{amsthm}
\usepackage{booktabs,bm}
\usepackage{algorithm}
\usepackage{algorithmic,multirow,enumitem}
\usepackage{amsmath, amsthm, amsfonts, amssymb}
\usepackage{url}
\usepackage{latexsym}
 
\usepackage{makecell}
\usepackage{hyperref}
\usepackage[T1]{fontenc}

\usepackage[utf8]{inputenc}

\usepackage{microtype}

\title{Relation Aware Semi-autoregressive Semantic Parsing for NL2SQL}

\author{

\makecell{Junyang Huang$^\S{^\dag}$,  Yongbo Wang$^\S$,  Yongliang Wang$^\S$, Yang Dong$^\S$, Yanghua Xiao$^\dag$} \\

\centerline{ 
$^\S$ Ant Group, 
$^\dag$ Fudan University
}\\

\centerline{ 
\tt{\{wyb269207,yongliang.wyl,doris.dy\}@antgroup.com},
}\\
\centerline{ 
\tt{\{jyhuang19,shawyh\}@fudan.edu.cn}}
}

\begin{document}
\newcommand{\MODN}{RaSaP}
\newcommand{\vect}[1]{\bm{{#1}}}
\maketitle
\begin{abstract}
Natural language to SQL (NL2SQL) aims to parse a natural language with a given database into a SQL query, which widely appears in practical Internet applications.
Jointly encode database schema and question utterance is a difficult but important task in NL2SQL.
One solution is to treat the input as a heterogeneous graph.
However, it failed to learn good word representation in question utterance.
Learning better word representation is important for constructing a well-designed NL2SQL system.
To solve the challenging task, we present a \textbf{R}elation \textbf{a}ware \textbf{S}emi-\textbf{a}utogressive Semantic \textbf{P}arsing (\MODN) ~framework, which is more adaptable for NL2SQL. 
It first learns relation embedding over the schema entities and question words with predefined schema relations with ELECTRA  and relation aware transformer layer as backbone.
Then we decode the query SQL with a semi-autoregressive parser and predefined SQL syntax.
From empirical results and case study, our model shows its effectiveness in learning better word representation in NL2SQL.

\end{abstract}


\section{Introduction}
Natural language to SQL (NL2SQL) task aims to parse a natural language with a given database into a SQL query.
NL2SQL is an important task in NLP and nowadays more and more NL2SQL systems are leveraged in web applications to provide question answering services.
Compared with most generic question answering tasks such as Knowledge Based Question Answering (KBQA) and Machine Reading Comprehension (MRC), NL2SQL has better interpretability since the SQL statement gives the reasoning path of the answer.
However, in NL2SQL, similar queries with different database schema leads to different SQL queries.
Thus, both database schema and word representation will affect the gold SQL query.

\begin{figure}[t] 
  \centering
  \includegraphics[scale=0.7]{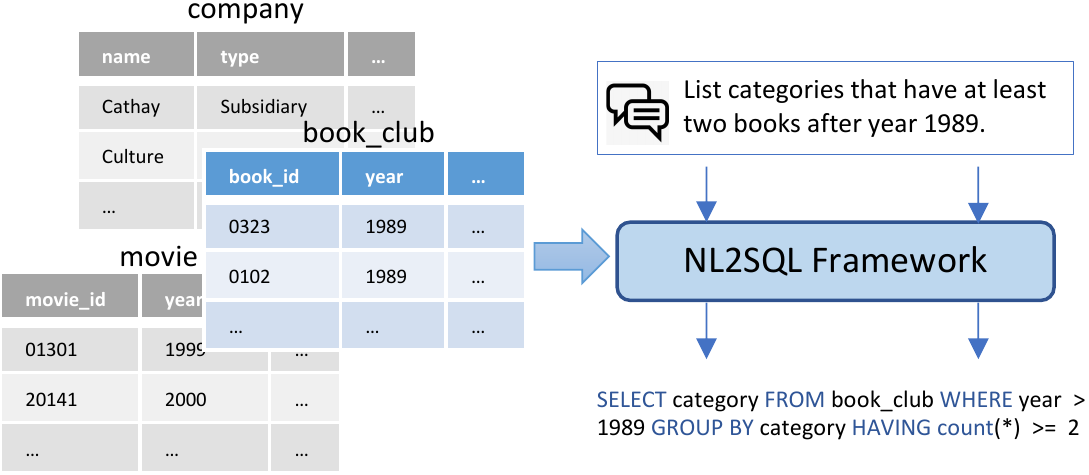}
  \caption{The illustration of an NL2SQL system. Given a question ``List categories that have at least two books after year 1989.'' and a database with several tables. The task of an NL2SQL framework is to parse the question into a SQL query ``SELECT category FROM book\_club WHERE year > 1989 GROUPBY category HAVING count(*) >= 2''}
  \label{fig:01}
\end{figure}

Building a system with the ability to jointly encode the database schema as well as the word representation is an important but difficult task in NL2SQL, which is critical to the final performance and user experience of an NL2SQL system.
Figure \ref{fig:01} is an example of NL2SQL.
Given a question ``List categories that have at least two books after year 1989.'' and a database with several tables.
The task of an NL2SQL framework is to parse the question into a SQL query ``SELECT category FROM book\_club WHERE year > 1989 GROUPBY category HAVING count(*) >= 2''.

Many solutions have been proposed to improve the logical and execution accuracy in NL2SQL.
Most of them focus on teaching the model learn better database schema representation.
Treating database schema as a heterogeneous graph \cite{cao2021lgesql,wang2020rat} to build the relations between question, column and table and learning the embedding of these relations to encode schema information into proposed models is a straightforward idea.
Enabling massive database schema information with pre-train models is another idea to enhance the ability of converting a question into a SQL query directly. 

\begin{figure*}[t] 
  \centering
  \includegraphics[scale=0.73]{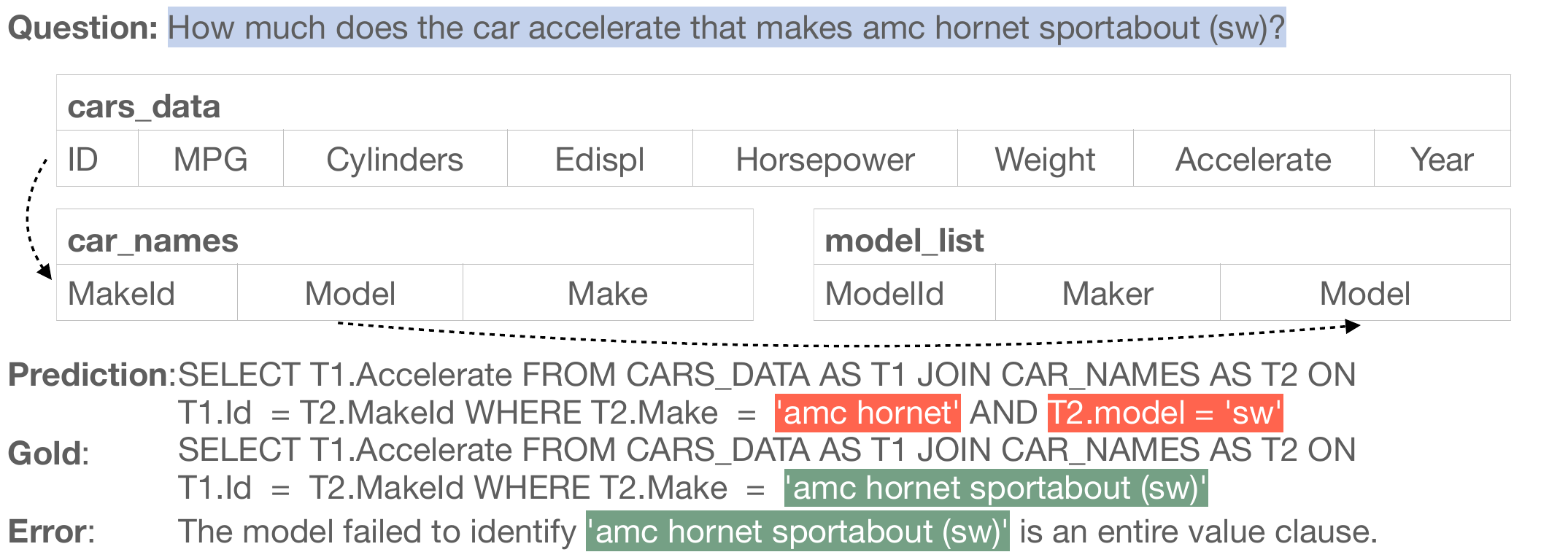}
  \caption{An illustration of a bad case. The model mistakenly take ``sportabout'' as a word that is irrelevant to SQL. Coincidentally, ``sw'' fits one value of the ``model'' column in the ``model\_list'' table. Therefore, ``amc hornet sportabout (sw)'' has been split into: ``amc hornet'' and ``sw'' which leads to the wrong SQL. }
  \label{fig:02}
\end{figure*}

However, despite database schema, word representation in questions is also very important.
For example, in Figure \ref{fig:02} the model failed to identiy ``amc hornet sportabout (sw)'' is an entire value clause.
The reason is the NL2SQL model mistakenly take ``sportabout'' as a word that is irrelevant to SQL. Coincidentally, ``sw'' fits one value of the ``model'' column in the ``model\_list'' table. 
Therefore, ``amc hornet sportabout (sw)'' has been split into: ``amc hornet'' and ``sw''. 
Actually ``sw'' is enclosed in parentheses and is semantically expressed as an abbreviation, but the model doesn't recognize it because the model can't capture and distinguish the latent semantic information of these word tokens.
In brief, model failed to learn good word representation.
Thus, learning better word representation is important for constructing a well-designed NL2SQL system.
We should discriminate which words are important in a question because some words in a question will confuse the model and some words are critical to the gold SQL.

%

To overcome the difficulties mentioned above, we propose a \textbf{R}elation \textbf{a}ware \textbf{S}emi-\textbf{a}utogressive Semantic \textbf{P}arsing (\MODN) ~framework, which is more adaptable for NL2SQL. 
We adopt the encoder-decoder framework to implement \MODN.
Thus, it consists of two parts, they are \MODN ~encoder and \MODN ~decoder.
For \MODN ~encoder, we uses relation-aware self-attention to learn relation embedding over the schema entities and question words with predefined schema relations. 
We then apply RAT-SQL\cite{wang2020rat} layer to jointly encode database schema and question utterance.
ELECTRA \cite{DBLP:journals/corr/abs-2003-10555} discriminator is used as the backbone of our encoder to learn better word representation and discriminate whether the token is important. 
For \MODN decoder, we follow the work of SmBop \cite{rubin2020smbop} and implement a semi-autoregressive bottom-up  parser that learns representations for meaningful semantic sub-programs at each step.
Empirical evaluation results demonstrate the the effectiveness of our model.
Specifically, we have made the following contributions:
\begin{itemize}
    \item \MODN ~learns better word representation in NL2SQL task while jointly learning word representation and database schema.
    \item \MODN ~discriminates the importance and correctness of a word token in the question utterance. 
    \item We demonstrated that discriminative pre-train model is effective in NL2SQL through our empirical experiments. 
\end{itemize}

\begin{table*}
  \centering
  \caption{The description of relation types present in the schema. An relation exists from source node $ x \in \mathcal{S} $ to target node $ y \in \mathcal{S} $ if the pair fulfills one of the descriptions listed in the table with the corresponding label. Otherwise, no edge exists from $x$ to $y$. }
  \begin{tabular}{*{6}c}  
  \toprule
Type of Node $x$ & Type of Node $y$ & Relation Label & Description  \\
    \midrule
    Q Token & Q Token & \textsc{Distance-d} & $y$ is the next $d_{th}$ question token of $x$.  \\
 \midrule
          \multirow{3}{*}{Column} & \multirow{3}{*}{Column}
   & \textsc{Same-Table}    & $x$ and $y$ belong to the same table. \\
 & & \textsc{Foreign-F} & $x$ is a foreign key for $y$. \\
 & & \textsc{Foreign-R} & $y$ is a foreign key for $x$. \\
 \midrule
 \multirow{2}{*}{Column} & \multirow{2}{*}{Table}
   & \textsc{Primary-Key-F}   & $x$ is the primary key of $y$. \\
 & & \textsc{Has-F}    & $x$ is a column of $y$ . \\
 \midrule
 \multirow{2}{*}{Table} & \multirow{2}{*}{Column}
   & \textsc{Primary-Key-R}   & $y$ is the primary key of $x$. \\
 & & \textsc{Has-R}    & $y$ is a column of $x$ . \\
 \midrule
 \multirow{3}{*}{Table} & \multirow{3}{*}{Table}
   & \textsc{Foreign-Tab-F}   & $x$ has a foreign key column in $y$. \\
 & & \textsc{Foreign-Tab-R}   & $y$ has a foreign key column in $x$. \\
 & & \textsc{Foreign-Tab-B}   & $x$ and $y$ have foreign keys in both directions. \\
    \midrule
              \multirow{3}{*}{Q Token} & \multirow{3}{*}{\shortstack{Column \\ or \\ Table}}
   & \textsc{No-Match}    & No overlapping between $x$ and $y$. \\
 & & \textsc{Partial-Match} & $x$ is a subsequence of the name of $y$ \\
 &  & \textsc{Exact-Match} & $x$ exactly matches the name of $y$ \\
 \midrule
 Q Token & Column & \textsc{Has-Value} & $x$ is part of one cell values of column $y$. \\
  \bottomrule
  \end{tabular}
  \label{tab:01}
\end{table*}


\section{Related Work}
\subsection{Encoding Queries for Text-to-SQL}

\citet{xu2017sqlnet} present ``column attention" strategy to tackle the joint encoding problem of the question and database schema.
\citet{guo2019towards} propose IRNET to solve the mismatch between intents expressed in natural language and the implementation details in SQL.
RAT-SQL \cite{wang2020rat} and DT-Fixup \cite{xu2020optimizing} encode relational structure in the database schema and a given question using relation-aware self-attention to combine global reasoning over the schema entities and question words with structured reasoning over predefined schema relations. 
LGESQL \cite{cao2021lgesql} constructs an edge-centric graph to explicitly considers the topological structure of edges with a line graph and proposes graph pruning to extract the golden schema items related to the question from the entire database schema graph.

\subsection{Tree Structured Decoding Strategy}

Capturing the target syntax as prior knowledge with abstract syntax tree \cite{yin2017syntactic,dong2018coarse,yin2021compositional} learns the underlying syntax of the target programming language.
The pervasiveness of tree structured decoding strategy is extensive, which are also widely used in solving math world problems.
\citet{wang2019rat} generate SQL in depth-first traversal order by using an LSTM to output a sequence of decoder actions with a grammar rule or column/table name.
\citet{rubin2020smbop} construct tokens at decoding step $t$ the top-K sub-trees of height $\leq t$ to improve decoding efficiency and learn representations for meaningful semantic sub-programs at each step.
\citet{xie2019goal} propose a model to generate an expression tree in a human-like goal-driven way by identifying and encoding its goal to achieve.
\citet{liu2019tree} employ tree structured decoder with BiLSTM  with a stack data structure to control the next generated token and when to stop decoding. 
Graph2Tree \cite{zhang2020graph} uses a graph transformer to learn the latent quantity representations from graphs, and a tree structure decoder to generate a solution expression tree.

\subsection{Data Augment for Text-to-SQL}
\citet{yu2020grappa} pre-trains GRAPPA on the  synthetic question-SQL pairs over high-quality tables via a synchronous context-free grammar to inject important structural properties commonly found in table semantic parsing.
GAP \cite{shi2021learning} jointly learns representations of natural language utterances and table schemas by leveraging generation models to generate pre-train data.
STRUG \cite{deng2021structure} proposes three tasks: column grounding, value grounding and column-value mapping and trains them using weak supervision without requiring complex SQL annotation.

\section{Task Definition}
\begin{figure*}[t] 
  \centering
  \includegraphics[scale=0.63]{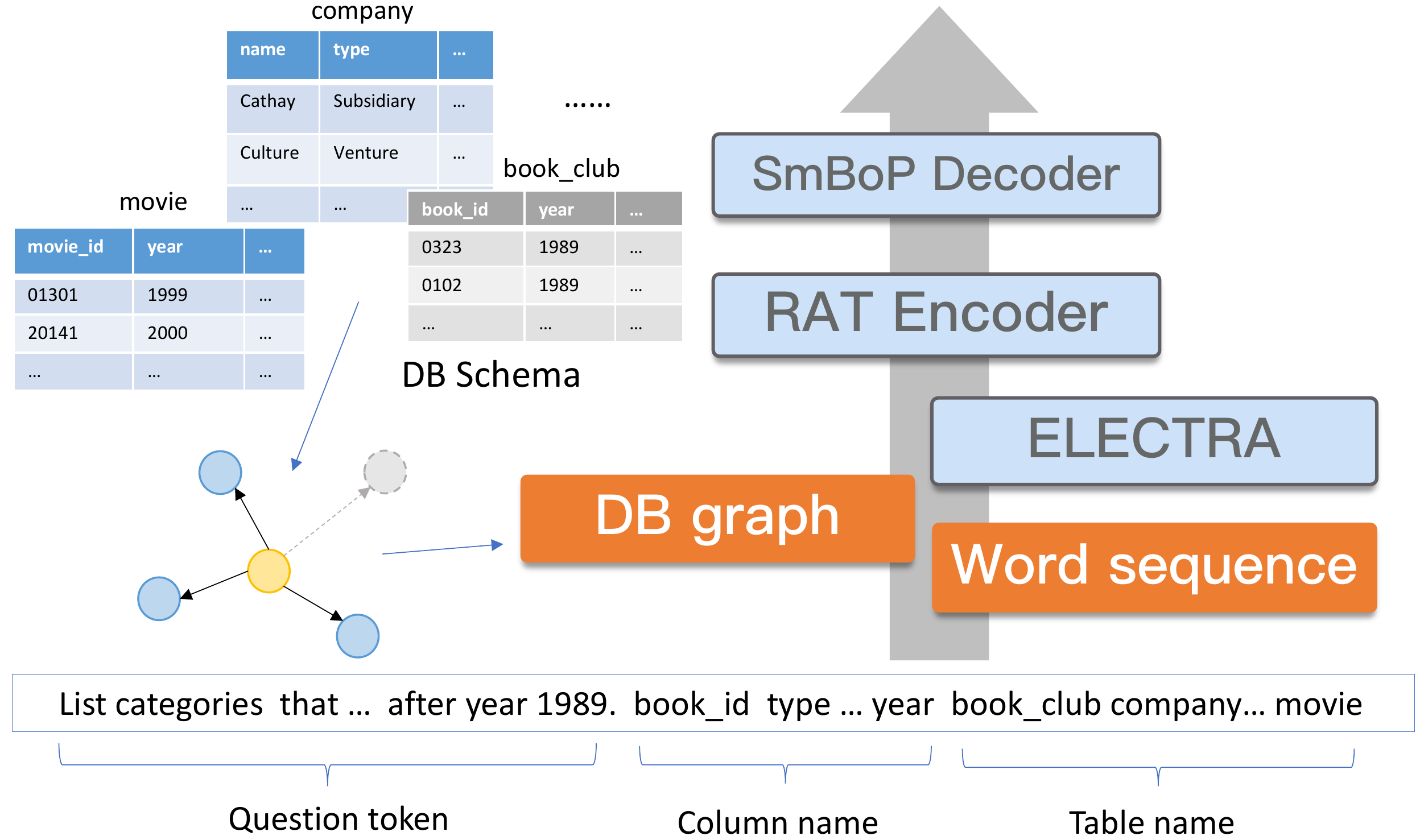}
  \caption{The framework of \MODN. We adopt the encoder-decoder framework to implement \MODN. Given the input as a graph $G$ and question $Q$, the RAT encoder $f_{enc}$ embeds it into joint representations $c_i$, $t_i$, $q_i$ for each column $c_i \in C$, table $t_i \in T$ , and question word $q \in Q$ respectively. The \MODN ~decoder $f_{dec}$ then uses them to compute a distribution  over the SQL programs.}
  \label{fig:03}
\end{figure*}

The input of our model is a natural language question and a corresponding database schema. 
The output is a SQL query $\mathcal{Z}$.
Specifically, given a question $\mathcal{Q}$ with $|\mathcal{Q}|$ tokens $\mathcal{Q}=[q_1,q_2,\dots,q_{|\mathcal{Q}|}]$ and the database schema $ \mathcal{S} = \mathcal{T} \cup \mathcal{C} $, the target is to generate the SQL $\mathcal{P}$. 
The schema $\mathcal{S}$ consists of tables  $\mathcal{T}=[t_1,t_2,\dots,t_{|\mathcal{T}|}]$ with $|\mathcal{T}|$ table names  and columns $\mathcal{C} = \{ c_1, c_2, \dots, c_{|\mathcal{C}|} \}$ with $|\mathcal{C}|$ column names.
Each column name $c_i$ contains words $c_i = \{c_{i,1}, c_{i,2}, \dots, c_{i, |c_i|}\}$ with length $|c_i|$ and each table name $t_i$ contains words $t_i = \{t_{i,1}, t_{i,2}, \dots, t_{i, |t_i|}\}$ with length $|t_i|$.
The database schema $ \mathcal{S} $ can be represent as a heterogeneous graph.
Its nodes consist of columns, tables and question each labeled with the words in its name (for columns, we constrain them with type $\tau \in \{ number, text\}$ to the label). 
Its edges are defined by the pre-existing database relations described in Table \ref{tab:01}.

\section{Methodology}

For modeling  \MODN , we adopt the encoder-decoder framework.
Given the input as a graph $G$ and question $Q$, the encoder $f_{enc}$ embeds it into joint representations $c_i$, $t_i$, $q_i$ for each column $c_i \in C$, table $t_i \in T$ , and question word $q \in Q$ respectively. 
The decoder $f_{dec}$ then uses them to compute a distribution  over the SQL programs and generated the final output query SQL $Z$.

\subsection{\MODN ~Encoder}

We implement our encoder $f_{enc}$ by following the state-of-the-art NLP literature RAT-SQL \cite{wang2019rat}. 
It first obtains a joint contextualized representation of the question and database schema. 
The initial representations $\vect{c^\text{{init}}}$,
$\vect{t^\text{{init}}}$ and $\vect{q^\text{{init}}}$ for every node of $G$ by retrieving the initial embedding from pre-train models.
The input $X$ is the set of all the node representations in $G$. Specifically, the question $Q$ is concatenated to a linearized form of the schema $S$.
Considering a set of inputs $X = \{\vect{x_i}\}_{i=1}^n$ where $\vect{x_i}\in \mathbb{R}^{d_x}$:
\begin{equation}
        X = (\vect{c}_{1}^\text{init}, \cdots, \vect{c}_{|\mathcal{C}|}^\text{init},
         \vect{t}_{1}^\text{init}, \cdots, \vect{t}_{|\mathcal{T}|}^\text{init},
         \vect{q}_{1}^\text{init}, \cdots, \vect{q}_{|\mathcal{Q}|}^\text{init}).
\end{equation}

Different with RAT-SQL, we use ELECTRA \cite{DBLP:journals/corr/abs-2003-10555} discriminator as the backbone of our encoder to initialize the initial representation $\vect{c_i^\text{{init}}}$, $\vect{t_i^\text{{init}}}$ and $\vect{q_i^\text{{init}}}$. 
Thus, tokens that correspond to a single schema constant are aggregated, which results in a final contextualized representation.
This contextualized representation  leads to better word representation and jointly learn the relations between the question and database schema.

Then the encoder uses relation aware self attention (RAT) to encode the database schema and question utterance.
RAT layer is quite same with the transformer \cite{vaswani2017attention} layer.
The difference between RAT layer and transformer layer is the self-attention is biased toward several relation embeddings noted as $\vect{\color{black} r_{ij}^K}$ and $\vect{\color{black} r_{ij}^V}$. 
$\vect{\color{black} r_{ij}^K}$ and $\vect{\color{black} r_{ij}^V}$ are learnable relation embedding representations.
We implement multi-head technique which consists of $H$ RAT layers with separate weight matrices in each layer.
Specifically, our encoder can be formulated as follows:

\begin{align}
    e_{ij}^{(h)} &= \frac{\vect{x_i} W_Q^{(h)} (\vect{x_j} W_K^{(h)} + \vect{\color{black} r_{ij}^K})^\top}{\sqrt{d_z
        / H}}   \\
    \alpha_{ij}^{(h)} &=\text{softmax}_{j} \bigl\{ e_{ij}^{(h)} \bigr\}, \\
    \vect{z}_i^{(h)} &= \sum_{j=1}^n \alpha_{ij}^{(h)} (\vect{x_j} W_V^{(h)} + \vect{\color{black} r_{ij}^V}), \\
     \vect{z}_i &= \text{Concat}\bigl(\vect{z}_i^{(1)}, \cdots, \vect{z}_i^{(H)}\bigr), \\
     \vect{\tilde{y}_i} &= \text{LayerNorm}(\vect{x_i} + \vect{z}_i), \\
    \vect{y_i} &= \text{LayerNorm}(\vect{\tilde{y}_i} + \text{FC}(\text{ReLU}(\text{FC}(\vect{\tilde{y}_i}))),
\end{align}
where $ W_Q^{(h)}, W_K^{(h)}, W_V^{(h)} \in \mathbb{R}^{d_x \times (d_x/H)} $.
FC and LayerNorm is a fully connected layer and layer normalization layer respectively.
Thus,
We refer the reader to \cite{wang2019rat} for details.

\subsection{\MODN ~Decoder}
We implement our decoder $f_{dec}$ by following the decoder of SmBop \cite{rubin2020smbop}.
The decoder will first compute the score of trees on the frontier. 
After scoring the frontier trees, the representation of trees will be scored.
Finally, the SQL constants and DB values will be generated during beam initialization (leaves). 
The parser returns the highest-scoring tree in $Z_t$ for step $t$. 
The decoder maintains a beam as follows at each decoding step:
\begin{equation}
    Z_t = ((z^{(t)}_1, \vect{z}^{(t)}_1), \dots, (z^{(t)}_K, \vect{z}^{(t)}_K)),
\end{equation}
where $z^{(t)}_i$ is the representation of the decoding tree, and $\vect{z}^{(t)}_i$ is its corresponding vector representation. 
Trees on the beams usually compose with each other (like a phrase). 
We contextualize tree representations on the beam using cross-attention. 
Specifically, we use self-attention \cite{vaswani2017attention} to give tree representations access to the input question:
\begin{equation}
    Z'_{t} \leftarrow \text{Attention}(Z_t, \vect{x}, \vect{x}), 
\end{equation}
where $(\vect{z}_{1}^{(t)}, \dots, \vect{z}_K^{(t)})$ 
are the queries from the tree representations, and the  $(\vect{x}_1,\dots,\vect{x}_{|x|})$ are the keys and values from input tokens respectively.
Next, let $e_\ell$ be an embedding for a unary or binary operation, and let $\vect{z}_i, \vect{z}_j$  be non-contextualized tree representations from the beam we are extending. 
We compute a new representation as follows:
\begin{align}
\vect{z}_\text{new} = \left\{\begin{matrix}
    \text{Transformer}(\vect{e}_\ell,\vect{z}_i) & \text{unary} \ell \\ 
    \text{Transformer}(\vect{e}_{\ell},\vect{z}_i,\vect{z}_j) & \text{binary } \ell \\
    \vect{z}_i & \ell = \textsc{Keep}
\end{matrix}\right.
\end{align}
where for the unary \textsc{Keep} operation, the representation of current step will be copied from the previous step. 
For $t+1$-high trees, we compute scores of them based on the frontier.
Frontier trees can be generated by applying a scoring function, where the score for a new tree $z_\text{new}$ generated by applying a unary rule $u$ on a tree $z_i$.
The score for a tree is generated by applying a binary rule $b$ on the trees $z_i, z_j$.
Specifically, they are defined as follows:
\begin{equation}
   s(z_\text{new}) = \vect{w}_u^\top FF_{U}([\textbf{z}_i; \textbf{z}_i']),
\end{equation}
\begin{equation}
s(z_\text{new}) = \vect{w}c_b^\top FF_{B}([\textbf{z}_i; \textbf{z}_i'; \textbf{z}_j; \textbf{z}_j']),
\end{equation}
where $FF_U$ and $FF_B$ are 2-hidden layer feed-forward layers with relu \cite{agarap2018deep} activation function.
After scoring the frontier, we generate a recursive vector representation for the top-$K$ trees.
In oder to guarantee the trees are generated with correct syntax and the generated SQL query is executable, we use semantic types to detect invalid rule applications and fix them. 

\section{Experiments}
\subsection{Dataset and Metrics}

We use the Spider dataset \cite{yu2018spider}, a challenging large-scale dataset for text-to-SQL parsing. 
Spider contains 8,659 training examples and 1034 evaluation examples.
Spider has become a common benchmark for evaluating NL2SQL systems because it includes complex SQL queries and a realistic zero-shot setup, where the database schema in test examples are different from training and evaluation examples.

As Spider makes the test set accessible only through an evaluation server, we perform our evaluations using the development set. 
We report our results using two metrics described in Spider leaderboard\footnote{\url{https://yale-lily.github.io/spider}}, they are: 1) exact set match accuracy without values and 2) execution accuracy with values, as well as divided by different difficulty levels.

\subsection{Training Details}
For the encoder $f_{enc}$, we encode the input utterance $\mathcal{Q}$ and the schema $\mathcal{S}$ with ELECTRA, consisting of $24$ Transformer layers, followed by another $H=8$ RAT-layers.
The dropout in RAT-layers are set as $0.2$.
For the decoder $f_{dec}$, we set the beam size as $K = 30$, and the number of decoding steps is $T = 9$ at inference time.
$9$ is chosen because the maximal tree depth on the development set is $9$.
The transformer used for tree representations has $1$ layer, $8$ heads, and with the dimension of $256$. 
The learning rate for language model and \MODN ~decoder is set to $3e-6$ and $0.000186$ respectively.
We train for $520$ epochs with batch size $12$ and gradient accumulation with $5$.

\subsection{Performance and Analysis}

We evaluate the accuracy of our model with the spider official evaluation script, which computes 1) exact set match without value (EM) and 2) execution with values (EXEC) since \MODN ~generates DB values.
For EM, the evaluation script will decompose each SQL into several clause instead of simply conducting string comparison between the predicted and gold SQL queries.
For EXEC, the evaluation script will evaluate whether the generated SQL can query the same value compared with the gold SQL.
The results are shown in Table \ref{tab:result}.
For model T5-3B+PICARD, since the parameter of the backbone model (3 billion) is approximately 4 times compared with other methods (770 million).
Thus we mainly compare our the models with the same parameter setups.

\begin{table}[t]
 \caption{Experiment results on the spider testset.}
 \begin{tabular}{lcc}
  \toprule
  Model & EM & EXEC \\
  \midrule
 RAT-SQL+STRUG & 68.4 & n/a \\
 RAT-SQL+GAP & 69.7  & n/a \\
 RAT-SQL+GRAPPA & 69.6  & n/a \\
 RAT-SQL+GP+GRAPPA & 69.8  & n/a \\
 BRIDGE+BERT(ensemble) & 67.5  & 68.3  \\
 \MODN+ELECTRA & \textbf{69.0} & \textbf{70.0}  \\
 \midrule
 T5-3B+PICARD & 71.9\% & 75.1\% \\
 \bottomrule
 \end{tabular}
 \label{tab:result}
\end{table}

\subsection{Case Study}

\begin{figure}[t] 
  \centering
  \includegraphics[scale=0.55]{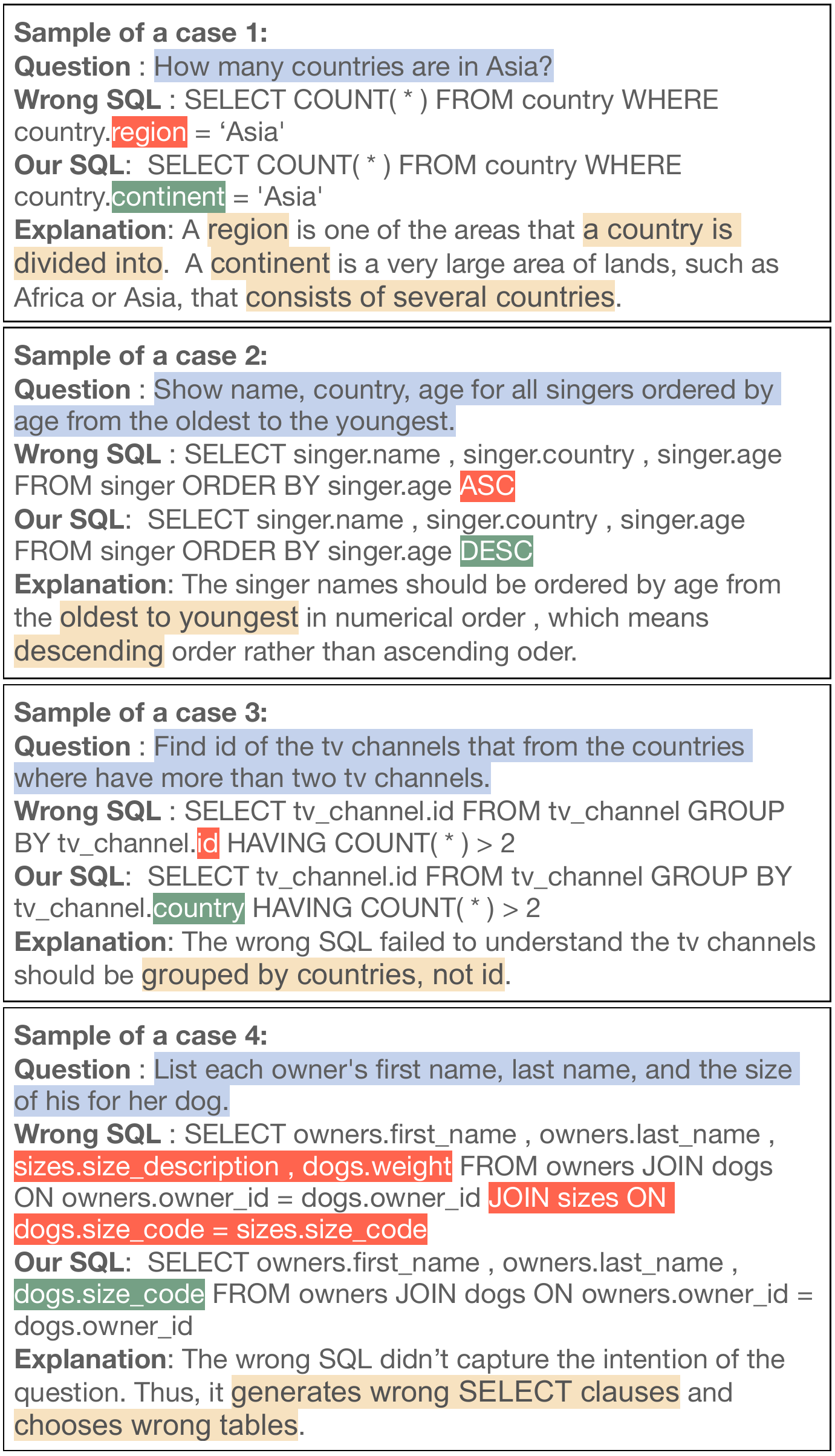}
  \caption{The case study of our model. For case 1, the where clause in gold SQL should be continent. For case 2, the singer names should be ordered by age with descending order rather than ascending order. For case 3, the tv channels should be grouped by countries, not id. For case 4, the intention of the query should be understand to generate the correct SQL.}
  \label{fig:cases}
\end{figure}

Since the test set used in evaluation script are unseen queries, we will make case study on the results of dev set. 
The cases are shown in fig \ref{fig:cases}.
We compare the result of our model with the result of SmBoP.
For case 1, a region is one of the areas that a country is divided into.  
A continent is a very large area of lands, such as Africa or Asia, that consists of several countries.
Thus, the where clause in gold SQL should be continent.
For case 2, the singer names should be ordered by age from the oldest to youngest in numerical order, which means descending order rather than ascending order.
So the SQL should sort the result with DESC rather than ASC.
For case 3, the wrong SQL failed to understand the tv channels should be grouped by countries, not id.
However, the result of our model is correct because our model can learn better word representation.
Therefore, our model can understand the question query better.
For case 4, the question is asking the size of the dog, first name and last name of the dog's owner.
However, the wrong SQL also failed to capture the intention of the question. 
Thus, it generates wrong SELECT clauses and chooses wrong tables.

\section{Conclusion}

In this paper, we present \MODN, an NL2SQL model to parse a natural language with a given database into a SQL query.
We illustrate that the proposed model can successfully learn better word representation in NL2SQL task while jointly learning word representation and database schema.
Moreover, it discriminates the importance and correctness of a word token in the question utterance.
Empirical results on spider dataset demonstrate the effectiveness of our model.
Thus, the proposed framework \MODN ~is effective in improving the performance of NL2SQL task.
In future work, we will explore novel approaches to improve the performance of the model on extra hard cases in spider.

\bibliography{custom}
\bibliographystyle{acl_natbib}

\appendix

\end{document}